\documentclass[11pt,letterpaper]{article}
\usepackage{emnlp2016}
\usepackage{times}
\usepackage{url}
\usepackage{latexsym}
\usepackage{color}
\usepackage{latexsym}
\usepackage{graphicx}
\usepackage{caption}
\usepackage[export]{adjustbox}
\usepackage{enumitem}
\usepackage{array}
\usepackage{float}
\usepackage{bbding}
\usepackage{multirow}
\label{my-label}
\emnlpfinalcopy

\title{Part of Speech Tagging for Code Switched Data }

\author{Fahad AlGhamdi, Giovanni Molina$^\ddagger$, Mona Diab, Thamar Solorio$^\ddagger$,\\ 
{\bf Abdelati Hawwari, Victor Soto $^\dagger$ \and Julia Hirschberg $^\dagger$} \\ \\
Department of Computer Science, The George Washington University \\
{\tt \{fghamdi,mtdiab,abhawwari\}@gwu.edu} \\
$^\ddagger${Department of Computer Science, University of Houston} \\
{\tt $^\ddagger${gemolinaramos@uh.edu solorio@cs.uh.edu}}\\
$^\dagger${Department of Computer Science, Columbia University} \\
{\tt $^\dagger${\{julia,vs2411\}@cs.columbia.edu}}}

\date{}

\begin{document}
\maketitle
\begin{abstract}
We address the problem of Part of Speech tagging (POS) in the context of linguistic code switching (CS). CS is the phenomenon where a speaker switches between two languages or variants of the same language within or across utterances, known as intra-sentential or inter-sentential CS, respectively. Processing CS data is especially challenging in intra-sentential data given state of the art monolingual NLP technology since such technology is geared toward the processing of one language at a time. In this paper we explore multiple strategies of applying state of the art POS taggers to CS data. We investigate the landscape in two CS language pairs,    
Spanish-English and Modern Standard Arabic-Arabic dialects. We compare the use of two POS taggers vs. a unified tagger trained on CS data. Our results show that applying a machine learning framework using two state of the art POS taggers achieves better performance compared to all other approaches that we investigate.


\end{abstract}

\section{Introduction}

Linguistic Code Switching (CS) is a phenomenon that occurs when multilingual speakers alternate between two or more languages or dialects. CS is noticeable in countries that have large immigrant groups, naturally leading to bilingualism. Typically people who code switch master two (or more) languages: a common first language (lang1) and another prevalent language as a second language (lang2). The languages could be completely distinct such as Mandarin and English, or Hindi and English, or they can be variants of one another such as in the case of Modern Standard Arabic (MSA) and Arabic regional dialects (e.g. Egyptian dialect-- EGY). CS is traditionally prevalent in spoken language but with the proliferation of social media such as Facebook, Instagram, and Twitter, CS is becoming ubiquitous in written modalities and genres \cite{HindiEnglshiCS,TheLanguageBook,CSANDMC}
CS can be observed in different linguistic levels of representation for different language pairs: phonological, morphological, lexical, syntactic, semantic, and discourse/pragmatic. It may occur within (intra-sentential) or across utterances (inter-sentential). For example, the following Arabic excerpt exhibits both lexical and syntactic CS. The speaker alternates between two variants of Arabic MSA and EGY.

\textbf{Arabic Intra-sentential CS:}\footnote{We use the Buckwalter encoding to present all the Arabic data in this paper: It is an ASCII only transliteration scheme, representing Arabic orthography strictly one-to-one} wlkn AjhztnA AljnA\}yp lAnhA m\$ xyAl Elmy lm tjd wlw mElwmp wAHdp.

\textbf{ English Translation}: Since our crime investigation departments are not dealing with science fiction, they did not find a single piece of information.

The speaker in the example switched from MSA to EGY dialect by using the word m\$/not which is an Egyptian negation particle, while s/he could have used the MSA word lyst/not. The span of the CS in this example is only one token, but it can be more than one example. Such divergence causes serious problems for automatic analysis.
CS poses serious challenges for language technologies, including parsing, Information Extraction (IE), Machine Translation (MT), Information Retrieval (IR), and others. The majority of these technologies are trained and exposed to one language at a time. However, performance of these technologies degrades sharply when exposed to CS data.

In this paper, we address the problem of part of Speech tagging (POS) for CS data on the intra-sentential level. POS tagging is the task where each word in  text is contextually labeled with grammatical labels such as, noun, verb, proposition, adjective, etc. 
We focus on two language pairs Spanish-English (SPA-ENG) and Modern Standard Arabic- and the Egyptian Arabic dialect (MSA-EGY). We use the same POS tag sets for both language pairs, the  Universal POS tagset \cite{UDPOS}. We examine various strategies to take advantage of the available monolingual resources for each language in the language pairs and compare against dedicated POS taggers trained on CS data for each of the language pairs. 
Our contributions are the following:
\begin{itemize}
\item We explore different strategies to leverage monolingual resources for POS tagging CS data.
\item We present the first empirical evaluation on POS tagging with two different language pairs. All of the previous work focused on a single language pair combination.
\end{itemize}

\section{Related Work }

Developing CS text processing NLP techniques for analyzing user generated content   as well as cater for needs of multilingual societies is vital \cite{HindiEnglshiCS}.
Recent research on POS for Hindi-English CS social media text conducted by \newcite{HindiEnglshiCS}, whereby social media text was proved to pose different challenges apart from CS, including transliteration, intentional and unintentional spelling differences, short and ungrammatical texts among others. Results indicated a significant improvement  where language detection as well as translation were automatically performed.  According to the study, accurate language detection as well as translation for social media CS text is important for POS tagging. However, they note that the juxtaposition of two monolingual POS taggers cannot solve POS tagging for CS text. \newcite{Jamatia} have also reported the challenge in POS tagging transliterated as well as CS social media text in Hindi English.  

\newcite{solorio} presents a machine learning based model that outperforms all baselines on SPA-ENG CS data. Their system  utilizes only a few heuristics in addition to the monolingual taggers. 

\newcite{sequierapos} introduces a ML-based approach with a number of new features. The new feature set considers the transliteration problem inherent in social media. Their system achieves an accuracy of 84\%. 

\newcite{jamatia2015} uses both a fine-grained and coarse-grained POS tag set in their study. They try to tackle the problem of POS tagging for English-Hindi Twitter and Facebook chat messages. They introduce a comparison between the performance of a combination of language specific taggers and a machine learning based approach that uses a range of different features. They conclude that the machine learning approach failed to outperform the language specific combination tagger.

\section{Approach}
The premise of this work is that monolingual resources should be helpful in POS tagging CS data. 
We adopt a supervised framework for our experimental set up. Supervised POS taggers are known to achieve the best performance, however they rely on significant amounts of training data. In this paper, we compare leveraging monolingual state of the art POS taggers using different strategies in what we call a COMBINED framework comparing it against using a single CS trained POS tagger identified as an INTEGRATED framework. 
We explore different strategies to investigate the optimal way of tackling POS tagging of CS data. To identify the underlying framework we prepend all COMBINED frameworks with the prefix COMB, and all the INTEGRATED versions with the prefix INT.
First we describe the monolingual POS taggers used in our set up. We consider the monolingual taggers to be our baseline systems.
\subsection{Monolingual POS Tagging systems}
We use a variant on the the publicly available  MADAMIRA tool \cite{Madamira} for the Arabic MSA-EGY pair. MADAMIRA is a supervised morphological disambiguator/tagger for Arabic text.  MADAMIRA extracts a wide variety of morphological and associated linguistic information from the input, including (among other things) detailed morphology and part-of-speech information, lemmas, fully-diacritized forms, and phrase-level information such as base phrase chunks and named entity tags. MADAMIRA is publicly available in two versions, an MSA version and EGY version. However, the publicly available version of MADAMIRA MSA is trained on newswire data (Penn Arabic Treebanks 1,2,3) \cite{Maamouri2004}, while MADAMIRA EGY is trained on Egyptian blog data which comprises a mix of MSA, EGY and CS data (MSA-EGY) from the LDC Egyptian Treebank parts 1-5 (ARZ1-5) \cite{Maamouri2012}. For our purposes, we need a relatively pure monolingual tagger per language variety (MSA or EGY), trained on informal genres for both MSA and EGY. Therefore, we retrained  a new version of MADAMIRA-MSA  strictly on pure MSA sentences identified in the EGY Treebank ARZ1-5. Likewise we created a MADAMIRA-EGY tagger trained specifically on the pure EGY sentences extracted from the same ARZ1-5 Treebank.\footnote{We are grateful to the MADAMIRA team for providing us with the MADAMIRA training code to carry out our experiments.}

For the SPA-ENG language pair we created models using the TreeTagger~\cite{schmid:94b} monolingual systems for Spanish and English respectively  as their performance has been shown to be competitive. Moreover, as pointed out in \cite{solorio} TreeTagger  has attractive features for our CS scenario. The data used to train TreeTagger for English was the Penn Treebank data~\cite{Marcus:1993}, sections 0-22. For the Spanish model, we used Ancora-ES \cite{tauleancora}. 
\subsection{Combined Experimental Conditions}
\begin{figure}[h]
  \centering
  
  \includegraphics[width=\columnwidth] {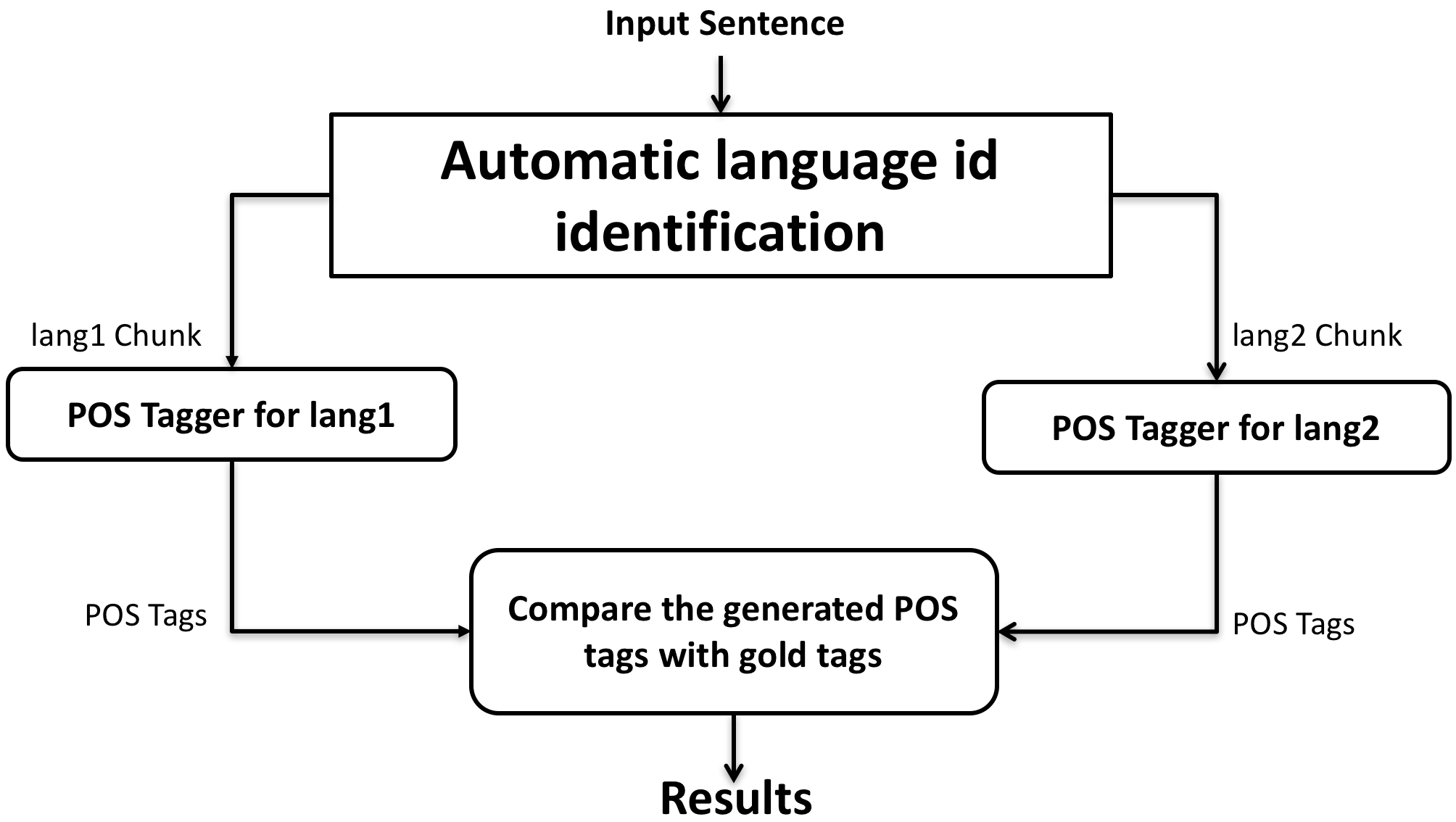}
   \caption{\label{fig:cond1} Graphic representation of the COMB1:LID-MonoLT approach.}
\end{figure}
\begin{figure}[h]
  \centering
  \small
  \includegraphics[width=\columnwidth] {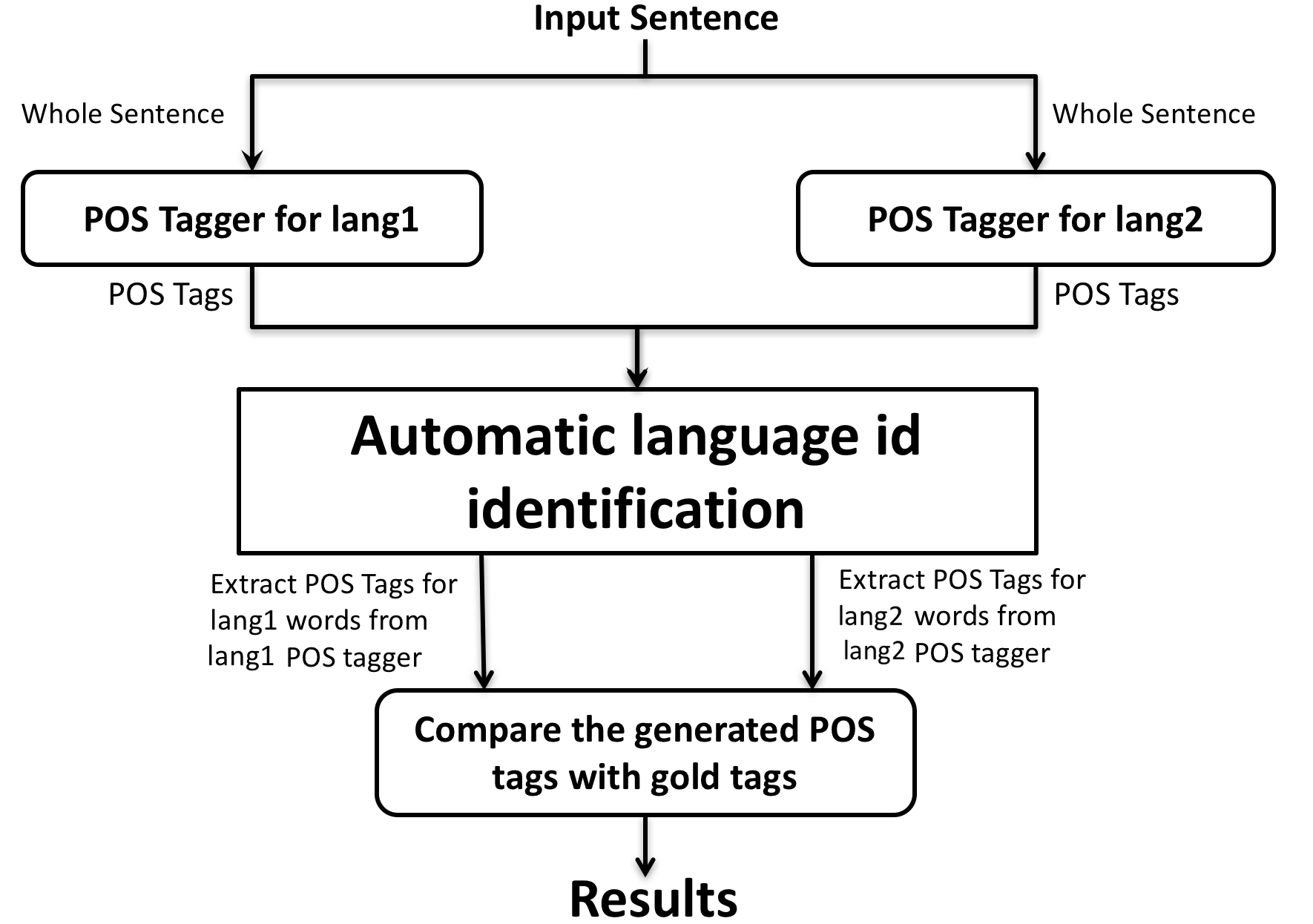}
   \caption{\label{fig:cond2} Graphic representation of the COMB2:MonoLT-LID approach.}
\end{figure}

\paragraph{COMB1:LID-MonoLT: Language identification followed by monolingual tagging}
Given a sentence, we apply a token level language identification process to the words in the sentence. The chunks of words identified as lang1 are processed by  the monolingual lang1 POS tagger and chunks of words identified as lang2 are processed by the monolingual lang2 POS tagger. Finally we integrate the POS tags from both monolingual taggers creating  the POS tag sequence for the sentence. Figure-\ref{fig:cond1} shows a diagram representing this approach for the MSA-EGY language pair. 
 For MSA-EGY, we used the Automatic Identification of Dialectal Arabic (AIDA2)  tool  \cite{AIDA2} to perform token level language identification for the EGY and MSA tokens in context. 
 It takes plain Arabic text in Arabic UTF8 encoding or Buckwalter encoding as input and outputs: 1) Class Identification (CI) of the input text to specify whether the tokens are MSA, EGY, as well as other information such as name entity, foreign word, or unknown labels per token. Furthermore, it provides the results with a confidence score; 2)Dialect Classification (DC) of the input text to specify whether it is Egyptian.
For SPA-ENG, we trained language models (LM) on English and Spanish data to assign Language IDs to each token in context. We trained  6-gram character language models using the SRILM Toolkit~\cite{stolcke}. The English language model was trained on the AFP section of the English GigaWord~\cite{EngGiga} while the Spanish language model was trained on the AFP section of the Spanish GigaWord~\cite{SpanGiga}.
\paragraph{COMB2:MonoLT-LID: Monolingual tagging then Language ID}
Similar to Condition COMB1, this experimental condition applies language ID in addition to monolingual tagging, however the order is reversed. In this condition we apply the two monolingual language specific POS taggers to the input CS sentence as a whole, then apply the language id component to the sentence, and then choose the POS tags assigned by the respective POS tagger per token as per its language id tag. The difference between this condition and condition 1 is that the monolingual POS tagger is processing an entire sentence rather than a chunk. It should be highlighted that all four monolingual POS taggers (ENG, SPA, MSA, EGY) are trained as sequence taggers expecting full sentence data as input. Figure- \ref{fig:cond2} shows a diagram representing this approach 
\paragraph{COMB3:MonoLT-Conf}
 In this condition, we apply separate taggers then use probability/confidence scores yielded by each tagger to choose which tagger to trust more per token. This condition necessitates that the taggers yield comparable confidence scores which is the case for the MADAMIRA-EGY and MADAMIRA-MSA pair, and the SPA-TreeTagger and EN-TreeTagger pair, respectively.
\paragraph{COMB4:MonoLT-SVM} 
In this condition, we combine results from  the monolingual taggers (baselines) and COMB3 into an ML framework such as SVM to decide which tag to choose from (MSA vs. EGY for example or SPA vs. ENG). By using an SVM classifier, we train a model on 10-fold cross-validation using the information generated by the monolingual POS taggers. The feature sets used for our model are the confidence scores and the POS tags generated by each tagger. Then, we evaluate our model on a held-out test set. 
\subsection{Integrated Experimental Conditions}
\paragraph{INT1:CSD}
In this condition, we train a supervised ML framework on exclusively code switched POS manually annotated data. In the case of  Arabic, we retrain a MADAMIRA model exclusively with the CS data  extracted from ARZ1-5 training data, yielding a MADAMIRA-CS model. For SPA-ENG, we trained a CS model using  TreeTagger. This provides consistency to the experimental set up but also allows us to compare the COMB and  INT approaches.  
 \paragraph{INT2:AllMonoData}
Similar to Condition INT1:CSD but changing the training data for each of the language pairs. Namely, we train a supervised ML framework on all the monolingual corpora that is POS manually annotated. For Arabic, we merge the training data from MSA and EGY, thereby creating a merged trained model. Likewise merging the Spanish and English corpora creating an integrated SPA-ENG model. The assumption is that the data in MSA is purely MSA and that in EGY is purely EGY. This condition yields an inter-sentential code switched training data set. None of the sentences reflect intra-sentential code switched data.   

\paragraph{INT3:AllMonoData+CSD}
Merging training data from conditions "INT1:CSD" and "INT2:AllMonoData" to train new taggers for CS POS tagging. 

\section{Evaluation}      
\subsection{Datasets}
For MSA-EGY, we use the LDC Egyptian Arabic Treebanks 1-5 (ARZ1-5) \cite{Maamouri2012}. The ARZ1-5 data is from the  discussion forums genre mostly in the Egyptian Arabic dialect (EGY). We refer to this data set as ARZ. Each part of the ARZ data set is  divided into train, test and dev sets. 
According to AIDA2 \cite{AIDA2}, 38\%  of the data exhibits CS. 
In this paper we combined the annotated test and dev set to form a new Code Switching Test data set, we refer to as ARZTest. The total number of words in the test data ARZTest is 20,464 tokens.

As mentioned earlier,  we created  new baseline POS tagging models for MADAMIRA-EGY, MADAMIRA-MSA and MADAMIRA-CS based on training data from ARZ1-5 training data portion. AIDA2 has a sentence level identification component that we used to identify the purity of the sentences from the training corpus ARZ1-5 training data. Specifically,  we used the AIDA2 identified EGY sentences for training the MADAMIRA-EGY  models, the MSA AIDA2 identified sentences for  training the MADAMIRA-MSA  models, and the CS identified AIDA2 sentences for training the MADAMIRA-CS models. 



For SPA-ENG data, We used two SPA-ENG CS data sets, one is the transcribed conversation used in the work by Solorio and Liu \cite{solorio}, referred to as Spanglish. The Spanglish data set has $\sim$8K tokens and was transcribed and annotated by the authors of that paper. While this is a small data set we include it in our work since it allows us to compare with previous work.

The second SPA-ENG CS data is the Bangor Miami corpus, referred to as Bangor. This corpus is also conversational speech involving a total of 84 speakers living in Miami, FL. In total, the corpus consists of 242,475 words of text from 35 hours of recorded conversation. Around 63\% of transcribed words are in English, 34\%  in Spanish and 3\%  in an indeterminate language. The transcriptions were carried out manually at the utterance level by a team of transcribers. They include beginning time and end time of utterance as well as language id for each word. Table~\ref{DatasetInfo} shows more details about the various data sets. 

\begin{table}[t]
\centering
\renewcommand{\arraystretch}{1}
\setlength{\tabcolsep}{0.3em}
\small
\begin{tabular}{|c|c|c|c|c|}
\hline
\textbf{Dataset}   & \textbf{\# sentences} & \textbf{\# Words} & \textbf{\# Types} & \textbf{\% CS} \\ \hline
\textbf{ARZ}       & 13,698              & 175,361             & 39,168              & 40.78\%        \\ \hline
\textbf{Spanglish} & 922                & 8,022               & 1,455                & 20.61\%        \\ \hline
\textbf{Bangor}    & 45,605              & 335,578             & 13,994              & 6.21\%        \\ \hline
\end{tabular}
\caption{Data set details.}
\label{DatasetInfo}
\end{table}

\begin{table}[t]
\centering
\renewcommand{\arraystretch}{1}
\setlength{\tabcolsep}{0.3em}
\small
\begin{tabular}{|c|c|c|}
\hline
\textbf{Dataset}   & \textbf{Train/Dev Tokens} & \textbf{Test Tokens} \\ \hline
\textbf{ARZ}       &  154,897             &     20,464           \\ \hline
\textbf{Spanglish} & 6,456                & 1,566   \\ \hline
\textbf{Bangor}    & 268,464              & 67,114   \\ \hline
\end{tabular}
\caption{Data set distribution.}
\label{DatasetInfo}
\end{table}

\subsection{Part of Speech Tagset}

The ARZ1-5 data set is manually annotated using the Buckwalter (BW) POS tag set. The BW POS tag set is considered one of the most popular Arabic POS tagsets. It gains its popularity from its use in the Penn Arabic Treebank (PATB)  \cite{Maamouri2004,NizarBuckwalter}. It can be used for tokenized and untokenized Arabic text. The tokenized tags that are used in the PATB are extracted from the untokenized tags. The number of untokenized tags is 485 tags and generated by BAMA \cite{Buckwalter2004}. Both tokenized and untokenized tags use the same 70 tags and sub-tags such as nominal suffix, ADJ, CONJ, DET, and, NSUFF \cite{Maamouri2009a} \cite{NizarBuckwalter}. Combining the sub-tags can form almost 170 morpheme sub-tags such NSUFF\_FEM\_SG. This is a very detailed tagset for our purposes and also for cross CS language pair comparison, i.e. in order to compare between trends in the MSA-EGY setting and the SPA-ENG setting. Accordingly, we map the BW tagset which is the output of the MADAMIRA tools to the universal tagset \cite{UDPOS}. We apply the mapping as follows: \textbf{1)} Personal, relative, demonstrative, interrogative, and indefinite pronouns are mapped to  Pronoun; \textbf{2)}Acronyms are mapped to Proper Nouns; \textbf{3)} Complementizers and adverbial clause introducers are mapped to Subordinating Conjunction; \textbf{4)}Main verbs (content verbs), copulas, participles, and some verb forms such as gerunds and infinitives are mapped to Verb; \textbf{5)}Adjectival, ordinal numerals and participles are mapped to Adjectives; \textbf{5)}Prepositions and postpositions are mapped to Adpositions; \textbf{6)}Interrogative, relative and demonstrative adverbs are mapped to Adverb; \textbf{7)}Tense, passive and Modal auxiliaries are mapped to Auxiliary Verb; \textbf{8)}Possessive determiners, demonstrative determiners, interrogative determiners, quantity/quantifier determiners, etc are mapped to Determiner; \textbf{9)} Noun and gerunds and infinitives are mapped to Noun; \textbf{10)}Negation particle, question particle, sentence modality, and indeclinable aspectual or tense particles are mapped to Particle.

The Bangor Miami corpus has also been automatically glossed and tagged with part-of-speech tags in the following manner: each word is automatically glossed using the Bangor Autoglosser \cite{don}.\footnote{\url{http://bangortalk.org.uk/autoglosser.php}}  Subsequently, transcripts were manually edited to fix incorrect glosses. 
For the experiments presented here, the corpus went through two edition/annotation stages. In the first stage, a number of changes were performed manually: a) those tokens ambiguously tagged with more than one POS tag were disambiguated (e.g. that.CONJ.[or].DET); b) ambiguous POS categories like ASV, AV and SV were disambiguated into either ADJ, NOUN, or VERB; c) for frequent tokens like \textit{so} and \textit{that}, their POS tags were hand-corrected; d) finally, mistranscribed terms which were originally labeled as Unknown were hand-corrected and given a correct POS tag.
The second stage consisted in mapping the Bangor corpus original POS tagset\footnote{\url{http://bangortalk.org.uk/docs/Miami_doc.pdf}} to the Universal POS tagset \cite{UDPOS}.\footnote{\url{http://universaldependencies.org/docs/u/pos/index.html}} After a careful examination of both tagsets, the following mapping was applied:
\textbf{1)} All those categories with an obvious match (like Nouns, Adjectives, Verbs, Pronouns, Determiners, Proper Nouns, Numbers, etc.) were automatically mapped; \textbf{2)} Exclamations and Intonational Markers were mapped to Interjections; \textbf{3)} As per the Universal POS tagset guidelines: Possessive Adjectives, Possessive Determiners, Interrogative Adjectives, Demonstrative Adjectives and Quantifying Adjectives were mapped to  Determiner; \textbf{4)} Those tokens tagged as Relatives, Interrogatives and Demonstratives (with no specification to whether they were Determiners, Adjectives or Pronouns) were manually labeled; \textbf{5)} All possessive markers, negation particles, and infinitive \textit{to} tokens were mapped to the PRT class; \textbf{6)} Conjunctions were mapped to Coordinating Conjunctions and Subordinating Conjunctions using word lists;
\begin{table}[H]
\centering
\renewcommand{\arraystretch}{1}
\setlength{\tabcolsep}{0.3em}
\small
\begin{tabular}{|c|c|c|}
\hline
\multicolumn{3}{|c|}{\textbf{MSA-EGY Baseline}}                  \\ \hline
Data set      & MADAMIRA-MSA       & MADAMIRA-EGY      \\ \hline
ARZTest & 77.23            & 72.22            \\ \hline
\multicolumn{3}{|c|}{\textbf{SPA-ENG Baseline}}         \\ \hline
Dataset      & TreeTagger SPA & TreeTagger ENG \\ \hline
Spanglish    & 44.61            & 75.87            \\ \hline
Bangor       & 45.95            & 64.05            \\ \hline
\end{tabular}
\caption{\small POS tagging accuracy (\%) for monolingual baseline taggers}
\label{t:baseline}
\end{table}

\textbf{7)} Finally, a subset of English Verbs were mapped to Auxiliary Verbs (could, should, might, may, will, shall, etc.).


\begin{table}[H]
\centering
\renewcommand{\arraystretch}{0}
\setlength{\tabcolsep}{0.1em}
\small
\renewcommand{\arraystretch}{1}
\begin{tabular}{|c|c|c|c|c|}
\hline 
\textbf{Approach}      & \textbf{Overall} & \textbf{CS}      & \textbf{MSA}     & \textbf{EGY}     \\ \hline
COMB1:LID-MonoLT     & 77.66           & 78.03          & 76.79          & 78.57          \\ \hline    
COMB2:MonoLT-LID       & 77.41           & 77.41          & 78.31          & 77.01          \\ \hline 
COMB3:MonoLT-Conf       & 76.66           & 77.89          & 76.79          & 76.11          \\ \hline
COMB4:MonoLT-SVM        & \textbf{90.56}  & \textbf{90.85} & \textbf{91.63} & \textbf{88.91} \\ \hline
INT1:CSD              & 83.89           & 82.03          & 82.48          & 83.26          \\ \hline
INT2:AllMonoData      & 87.86           & 87.92           & 86.82           & 86\%              \\ \hline
\begin{tabular}[c]{@{}c@{}}INT3:AllMonoData+CSD\end{tabular} & 89.36           & 88.12           & 85.12           & 87              \\ \hline
\end{tabular}
\caption{Accuracy (\%) Results for ARZTest Dataset}
\label{ARZ}
\end{table}
\begin{table}[ht]
\centering
\renewcommand{\arraystretch}{0}
\setlength{\tabcolsep}{0.1em}
\small
\renewcommand{\arraystretch}{1}
\begin{tabular}{|c|c|c|c|c|}
\hline
\textbf{Approach}      & \textbf{Overall} & \textbf{CS} & \textbf{ENG} & \textbf{SPA} \\ \hline
COMB1:LID-MonoLT        & 68.35          & 71.11          & 66.36          & 76.02\%          \\ \hline
COMB2:MonoLT-LID        & 65.51          & 69.66          & 64.44          & 71.32\%          \\ \hline
COMB3:MonoLT-Conf       & 68.25          & 68.21          & 71.93          & 65.03          \\ \hline
COMB4:MonoLT-SVM        & \textbf{96.31} & \textbf{95.39} & \textbf{96.37} & \textbf{96.60} \\ \hline
INT1:CSD              & 95.28          & 94.41          & 94.41          & 95.15          \\ \hline
INT2:AllMonoData      & 78.57          & 78.62          & 81.85          & 76.53\%          \\ \hline
\begin{tabular}[c]{@{}c@{}}INT3:AllMonoData+CSD\end{tabular} & 91.04          & 89.59          & 92.00          & 89.48          \\ \hline
\end{tabular}
\caption{Accuracy (\%) Results for Bangor Corpus}
\label{Bangor}
\end{table}

\begin{table}[H]
\renewcommand{\arraystretch}{1}
\setlength{\tabcolsep}{0.1em}
\small
\centering
\begin{tabular}{|c|c|c|c|c|}
\hline
Approach               & \textbf{Overall} & \textbf{CS}      & \textbf{ENG} & \textbf{SPA} \\ \hline
COMB1:LID-MonoLT        & 78.73          & 77.81          & 80.18          & 73.99          \\ \hline
COMB2:MonoLT-LID        & 73.52          & 73.80          & 73.60          & 71.57          \\ \hline
COMB3:MonoLT-Conf       & 77.39          & 76.11          & 80.20          & 65.43          \\ \hline
COMB4:MonoLT-SVM        & \textbf{90.61} & \textbf{89.43} & \textbf{93.61} & \textbf{87.96} \\ \hline
INT1:CSD              & 82.95          & 83.03          & 85.95          & 77.26          \\ \hline
INT2:AllMonoData      & 84.55          & 84.84          & 88.50          & 76.59          \\ \hline
\begin{tabular}[c]{@{}c@{}}INT3:AllMonoData+CSD\end{tabular} & 85.06          & 84.70          & 90.15          & 76.59          \\ \hline
\end{tabular}
\caption{Accuracy (\%) Results for Spanglish Corpus }
\label{Spanglish}
\end{table}

To evaluate the performance of our approaches we report the accuracy of each condition by comparing the output POS tags generated from each condition against the available gold POS tags for each data set.  
Also, we compare the accuracy of our approaches for each language pair to its corresponding monolingual tagger baseline. We consistently apply the different experimental conditions on the same test set per language pair: for MSA-EGY we report results on ARZTest, and for SPA-ENG, we report results on two test sets: Spanglish and Bangor. 
\paragraph{Baseline Results}
The baseline performance is the POS tagging accuracy of the monolingual models with no special training for CS data. Since we have four monolingual models, we consider four baselines. If CS data do not pose any particular challenge to monolingual POS taggers, then we shouldn't expect a major degradation in performance. Table \ref{t:baseline} shows the performance of the 4 different baseline POS tagging systems on the test data. For Arabic, MSA monolingual performance for MADAMIRA-MSA, when tested on monolingual MSA test data, is around $\sim$97\% accuracy, and for MADAMIRA-EGY when tested in monolingual EGY data it is $\sim$93\%. We note here that the presence of CS data in the ARZ test data causes these systems to degrade significantly in performance (77\% and 72\% accuracy, respectively). For SPA-ENG, state of the art monolingual models achieve an accuracy of $\sim$96\% and $\sim$93 on monolingual English and monolingual Spanish data sets, respectively. It is then clear that CS data poses serious challenges to monolingual technology. Other prior work has also reported similar drops in performance because of having mixed language data.

\subsection{Results}
Table~\ref{ARZ}, Table~\ref{Bangor} and Table~\ref{Spanglish} show the results of all our experimental conditions. For all language pairs, we report four results, the accuracy results for only lang1 sentences, lang2 sentences, CS sentences, and all sentences. For example for MSA-EGY, we extract the MSA, EGY, and CS sentences, respectively, from each experimental setup to report the breakdown of the performance of the condition on the specific set, i.e. We calculate the accuracy for the MSA, EGY, CS, and All (MSA+EGY+CS) sentences.
For MSA-EGY the highest accuracy is 90.56\%. It is achieved  when we apply condition "COMB4:MonoLT-SVM". All the INT conditions outperform the COMB conditions except for COMB4:MonoLT-SVM. Among the COMB conditions, we note that the MonoLT-SVM is the best COMBINED condition. 
In the SPA-ENG, the highest accuracies are achieved when we apply condition "COMB4:MonoLT-SVM". This finding of having the best POS tagging results when using the monolingual taggers output to train a machine learning algorithm confirms the concluded results in \cite{solorio}. Our contribution is that we replicate these results using a unified POS tagging scheme using an additional, much larger data set.
The lowest accuracy in SPA-ENG is when we apply "COMB2:MonoLT-LID" condition. The accuracy reaches $\sim$73\% in the Spanglish dataset and $\sim$65\% for the Bangor corpus. In this language pair the INT conditions outperform all the COMB except the one that uses the stack-based approach (COMB4:MonoLT-SVM). 
It is interesting that we observe the same trends across both language pairs. 
\section{Discussion}
\paragraph{Combined conditions}
For MSA-EGY, all the combined experimental conditions outperform the baselines. Among the combined conditions we note that applying language identification then applying language specific monolingual taggers yields worse results than applying the taggers on the input sentence then assigning tags as per the language ID tool. This is expected due to the fact that the taggers are expecting well formed sentences on input. Applying condition LID-MonoLT forces the taggers to tag chunks as opposed to sentences thereby leading to degraded performance. 
For the first two conditions COMB1:LID-MonoLT and COMB2:MonoLT-LID, the performance increases slightly from 77.41\% to 77.66\%. 
The results for MonoLT-SVM are the highest for the combined conditions for MSA-EGY. The worst results are for condition MonoLT-Conf. This might be relegated to the quality of the confidence values produced by the monolingual taggers, i.e. not being very indicative of the confidence scores for the tags chosen. 
For the SPA-ENG language pair, almost all the accuracies achieved by the combined conditions are higher than the Spanglish data set's baselines. The only combined condition that is lower than the baselines' of the Spanglish data set is "COMB2:MonoLT-LID" condition, where the accuracy is 73.52\% compared to the monolingual English tagger that reached a baseline performance of 75.87\%. This difference can be attributed to mistakes in the  automated language identification that cause the wrong tagger to be chosen. 

If we consider all accuracy results of the combined conditions for the Bangor corpus and its baselines, we see that the boosts in accuracy are of at least 2\%. It is quite noteworthy that the trends seem to be the same between the two language pairs. Both language pairs achieve the highest performance with MonoLT-SVM and worse results with MonoLT-Conf. The gains in performance from using a learning algorithm are likely due to the fact that the learner is taking advantage of both monolingual tagger outputs and is able to go beyond the available tags for cases where there are errors in both. This result is also consistent with findings in the Spanglish dataset by \cite{solorio}. The weaknesses of the MonoLT-Conf approach probably come from the fact that if the monolingual taggers are weak, their confidence scores are equally unreliable.

However the results are switched between conditions LID-MonoLT (condition 1) and MonoLT-LID (condition 2) for the two language pairs. Condition 1 outperforms condition 2 for MSA-EGY while we see the opposite for SPA-ENG. This is an indication of the quality and robustness of the underlying strength of the SPA and ENG monolingual taggers, they can handle chunks more robustly compared to the Arabic taggers. It is worth noting that the underlying Language id component for Arabic, AIDA2, achieves a very high accuracy on token and sentence level dialect id, F1 92.9\% for token identification, and an F1 of 90.2\% on sentence level dialect identification. Also compared to manual annotation on the TestDev set for dialect identification, we note an inter-annotator agreement of 93\% between human annotation and AIDA2. 

All COMB conditions use either out of context or in context chunks as an input for the monolingual taggers. We believe that the out of context chunks especially in the MSA-DA language pair contributed heavily in the noncompetitive results yielded.
    
\paragraph{Integrated conditions}
The rather simple idea of throwing the monolingual data together to train a model to label mixed data turned out to reach surprisingly good performance across both language pairs.
In general, except the "COMB4:MonoLT-SVM" condition all the INT conditions outperformed the COMB conditions and in turn the baselines for the MSA-EGY language pair. For this language pair we note that adding more data helps, INT2:AllMonoData outperforms INT1:CSD, but combining the two conditions as training data, we note that INT3:AllMonoData+CSD outperforms the other INT conditions. Applying the INT conditions on only the CS sentences yields the highest accuracy compared to the other sentences types. 
For SPA-ENG, the worse INT condition is INT2:AllMonoData for Bangor (accuracy 78.57\%) and INT1:CSD for Spanglish (accuracy 82.95\%), compared to the best performing condition for both SPA-ENG data sets, Spanglish (accuracy of 90.61\%) and Bangor (accuracy 96.31\%). The largest difference is in the Bangor corpus itself and this gap in performance could be due to a higher domain mismatch with the monolingual data used to train the tagger. Another notable difference between the two language pairs is the significant jump in performance for the Bangor corpus from the first three COMB conditions from 68.35\% to 96.31\%. While we observe a similar jump for the Spanglish corpus, the gap is much larger for the Bangor corpus. Here again, we believe the major factor is a larger mismatch with the training corpus for the monolingual taggers.
It should be highlighted that even though the language pairs are very different, there are some similar trends between the two combinations. COMB4:MonoLT-SVM is the best among combined conditions and INT conditions for the two language pairs. Moreover, except for the COMB4:MonoLT-SVM condition, all the INT conditions outperform combined conditions across the board. 
The percentage of the CS sentences in the ARZ dataset is $\sim$51\%. Moreover, MSA and EGY share a significant number of homographs some of which are cognates but many of which are not. This could be contrasted to the SPA-ENG case where the homograph overlap is quite limited. Adding the CSD to the monolingual corpora in the INT3:AllMonoData-CSD condition for MSA-EGY improves performance (1.5\% absolute increase in accuracy) allowing for more discriminatory data to be included comparing to the other INT conditions, while the results are not consistent across the SPA-ENG data sets.
In general, our results point to an inverse correlation between language similarity and the challenge to adapt monolingual taggers to a language combination. MSA-EGY has higher average baseline performance than SPA-ENG 
and all approaches outperform by a large margin those baseline results. In contrast, the average baseline performance for SPA-ENG is lower and the improvements gained by the approaches explored have different degrees of success. Additional studies are needed to further explore the validity of this finding.
\section{Conclusions}
We presented a detailed study of various strategies for POS tagging of CS data in two language pairs. The results indicate that depending on the language pair and the distance between them there are varying degrees of need for annotated code switched data in the training phase of the process. Languages that share a significant amount of homographs when code switched will benefit from more code switched data at training time, while languages that are farther apart such as Spanish and English, when code switched, benefit more from having larger monolingual data mixed. 
All COMB conditions use either out of context or in context chunks as an input for the monolingual taggers. We believe that out of context chunks especially in the MSA-DA language pair contributed heavily in the noncompetitive results that we got for the COMB conditions. Therefore, our plan for the future work that process the out of context chunks to provide a meaningful context to the monolingual taggers. Also, we plan to extend our feature set used in the COMB4:MonoLT-SVM condition to include Brown Clustering, Word2Vec, and Deep learning based features.    





\bibliography{pos}
\bibliographystyle{emnlp2016}



\end{document}